# Challenges in Credit Assignment for Multi-Agent Reinforcement Learning in Open Agent Systems

Alireza Saleh Abadi
School of Computing
University of Nebraska-Lincoln
asalehabadi2@unl.edu

Leen-Kiat Soh
School of Computing
University of Nebraska-Lincoln
lksoh@unl.edu

## ABSTRACT

In the rapidly evolving field of multi-agent reinforcement learning (MARL), understanding the dynamics of open systems is crucial. Openness in MARL refers to the dynamic nature of agent populations, tasks, and agent types within a system. Specifically, there are three types of openness as reported in (Eck et al. 2023) [2]: agent openness, where agents can enter or leave the system at any time; task openness, where new tasks emerge, and existing ones evolve or disappear; and type openness, where the capabilities and behaviors of agents change over time. This report provides a conceptual and empirical review, focusing on the interplay between openness and the credit assignment problem (CAP). CAP involves determining the contribution of individual agents to the overall system performance, a task that becomes increasingly complex in open environments. Traditional credit assignment (CA) methods often assume static agent populations, fixed and pre-defined tasks, and stationary types, making them inadequate for open systems. We first conduct a conceptual analysis, introducing new sub-categories of openness to detail how events like agent turnover or task cancellation break the assumptions of environmental stationarity and fixed team composition that underpin existing CAP methods. We then present an empirical study using representative temporal and structural algorithms in an open environment. The results demonstrate that openness directly causes credit misattribution, evidenced by unstable loss functions and significant performance degradation. The contributions are threefold: (1) a conceptual analysis introducing new sub-categories of openness that challenge the foundations of traditional credit assignment methods; (2) an empirical evaluation discussing how openness may cause credit misattribution and degraded performance; and (3) the identification of failure points and research gaps, motivating the development of new CAP methods tailored to open systems.

## 1. Introduction

Multi-agent reinforcement learning (MARL) enables agents to learn cooperative or competitive behaviors through repeated interactions with their environment. While many MARL methods assume closed systems with fixed agents and tasks, real-world applications are inherently open. Agents may enter or leave, tasks may appear or disappear, and agent types may change mid-episode. This paper focuses on how such openness impacts a core challenge in MARL: the credit assignment problem (CAP).

CAP refers to determining which decisions, actions, components, or agents contributed to observed outcomes. CAP is typically divided into two categories: *temporal credit assignment (TCA)*, which links actions to delayed rewards over time, and *structural credit assignment (SCA)*, which attributes outcomes to specific components such as agents or network modules [24]. When credit is misassigned, it can lead to slower learning, sub-optimal policies, unstable reward propagation, and in severe cases, catastrophic unlearning [4, 19].

Openness introduces further complexity by violating the fixed assumptions on which many CAP methods rely. As defined by Eck et al. [2], there are three types of openness: agent openness, where agents may join or leave; task openness, where tasks may appear or disappear; and type openness, where an agent's capabilities, preferences, or goals may change over time. Each form changes the structure or dynamics of the environment, making temporal and structural credit assignment (CA) significantly more difficult.

This work investigates how openness affects the CAP through conceptual and empirical analyses. Our conceptual analysis is grounded in established literature and informed by our own analysis of how openness breaks or complicates assumptions of existing methods. For the empirical analyses, we evaluate two representative algorithms, i.e., Deep Q-Network (DQN) [11] for TCA, and Multi-Agent PPO (MAPPO) [34] for SCA, respectively. Each method is adapted to operate in an environment with openness. To measure the impact of openness on CAP, we track key indicators including convergence speed, adaptability, and reward per episode caused by openness. These metrics allow us to quantify learning complications produced by poor CA under different forms of openness.

In particular, the main contributions of this work are as follows:

- We identify and categorize how specific openness events directly violate the stationarity and

compositional assumptions that existing TCA and SCA methods rely on.

- We conduct experiments using representative algorithms for TCA and SCA in an open environment. Our results provide empirical evidence that openness causes significant performance degradation and learning instability, linking these failures to credit misattribution.

- Based on our analyses, we discuss limitations in current approaches regarding openness and outline possible research gaps, establishing the need for openness-aware CAP methods and more comprehensive evaluation benchmarks.

In the following, Section 2 provides background on reinforcement learning (RL), MARL, and the challenges of coordination and credit assignment, an introduction of agent openness, task openness, and type openness, and a detailed discussion on the CAP in MARL. Section 3 presents a conceptual analysis of how each type of openness affects CAP, and Section 4 is our empirical analysis to evaluate these effects in a wildfire domain [37] using DQN and MAPPO algorithms for TCA and SCA. Section 5 outlines remaining research gaps and directions for openness-robust CAP methods. Section 6 concludes by summarizing this review's insights and contributions.

## 2. Background

This section introduces RL and its extension to multi-agent systems (MARL). It then discusses openness in MARL, highlighting agent openness, task openness, and type openness. Finally, it presents the CAP.

### 2.1 Single and Multi-Agent RL

RL is a sequential decision-making framework where an agent learns to maximize cumulative reward through interactions with its environment. The environment is typically modeled as a Markov Decision Process (MDP) [17], defined by a tuple ⟨S, A, P, R, γ⟩. At each timestep, the agent observes a state $s_t \in S$, selects an action $a_t \in A$, receives a reward $r_t = R(s_t, a_t)$, and transitions to a new state $s_{t+1} \sim P(\cdot | s_t, a_t)$ [6, 24].

The agent learns a policy $\pi(a|s)$ that maximizes the expected return $\mathbb{E}[\sum_{k=0}^{\infty} \gamma^k r_{t+k}]$. Central to many RL algorithms are the Bellman equations [6, 24], which recursively define the value of states and state-action pairs under a given policy, as shown in Equation 1 below:

$$V^\pi(s) = E_{a \sim \pi}[R(s,a) + \gamma V^\pi(s')] \quad (1)$$

MARL extends RL to environments involving multiple agents, each with its own policy and possibly its own reward signal. These agents interact in a shared environment, leading to additional challenges such as non-stationarity, coordination, and joint decision spaces. MARL is often formalized as a Markov Game [10] $\langle I, S, \{A_i\}, P, \{R_i\}, \gamma \rangle$, where each agent $i \in I$ selects actions $a_i \in A_i$ and receives a reward $R_i$ [35].

The CAP is a pivotal challenge in RL, fundamentally asking how an agent determines which past actions or components led to an observed reward [24]. In MARL, such reasoning become particularly complex due to shared rewards and intricate agent interactions, necessitating robust mechanisms to accurately assign credit.

### 2.2 Credit Assignment Problem (CAP)

The CAP concerns determining which agents, actions, or components contributed to an observed outcome. In MARL, shared rewards and delayed feedback make this difficult. Most algorithms assume fixed agents and tasks, making them brittle under openness. This problem is broadly categorized into two types: *TCA* and *SCA*.

*Temporal Credit Assignment (TCA)*. TCA is typically addressed using methods like temporal difference (TD) learning [23] or eligibility traces [21], which propagate delayed rewards. For instance, in Q-learning [31], a popular method, the Q-value for a state-action pair $(s_t, a_t)$ is updated towards a target that includes the immediate reward and the discounted maximum future Q-value, as shown in Equation 2:

$$Q(s_t, a_t) \leftarrow Q(s_t, a_t) + \alpha \left[ r_t + \gamma \max_{a'} Q(s_{t+1}, a') - Q(s_t, a_t) \right] \quad (2)$$

In Equation 2, the current estimate $Q(s_t, a_t)$ is updated using the learning rate $\alpha$, the immediate reward $r_t$, the discount factor $\gamma$, and the estimated maximum future value $\max_{a'} Q(s_{t+1}, a')$, which represents the best possible future reward from the next state. Deep Q-Networks (DQN) [11] extend this concept by using neural networks to approximate the Q-function, enabling learning in high-dimensional state spaces.

*Structural Credit Assignment (SCA)*. SCA, on the other hand, deals with attributing outcomes to specific agents or components within a system. Policy gradient methods are commonly employed here where the policy parameters are adjusted in the direction that increases the expected return. A general form of the policy gradient can be seen as optimizing for each agent $i$ as shown in Equation 3:

$$\nabla J(\theta_i) = \mathbb{E}_{\pi_{\theta_i}} \left[ \nabla_{\theta_i} \log \pi_{\theta_i}(a_i|s) A(s,a) \right] \quad (3)$$

In Equation 3, $\theta_i$ represents the parameters of agent $i$'s policy, $\pi_{\theta_i}(a_i|s)$ is the probability of taking action $a_i$ in state $s$ under policy $\theta_i$, and $A(s,a)$ is an advantage function that quantifies how much better an action is than the average for a given state [25]. Algorithms like Multi-Agent Proximal Policy Optimization (MAPPO)[34] leverage centralized critics and decentralized actors to estimate these advantage functions more effectively in MARL settings, helping to disentangle individual agent contributions from joint rewards.

*Assumptions*. Both types of CA become significantly more complex and challenging under the conditions of *openness* in MARL, as assumptions about stationary dynamics, known agent identities, and fixed coordination structures are often not being met.

More specifically, we see that temporal and SCA methods rely on several key assumptions to ensure accurate reward propagation and credit attribution. When these assumptions do not hold due to openness, standard CA techniques become unreliable, requiring a deeper analysis of their limitations. As reported in literature, the key assumptions are: (1) *Stationary Environment*, where the environment is assumed to remain stable throughout an episode, meaning that tasks, rewards, and agent interactions do not change [6, 32]; (2) *Fixed Set of Agents*, where the number of agents remains constant, ensuring that credit can be consistently assigned to known participants [20]; and (3) *Stable Reward Function*, where rewards are assumed to be consistently mapped to actions, predefined tasks, and agent behaviors [6]. In addition, we also identify two critical assumptions: (4) *Markov Property* where future states and rewards depend only on the current state and action, allowing for predictable credit propagation [6, 24], and (5) *Consistent Action-Outcome Mapping* where the same action in the same state is expected to produce similar outcomes [5, 10]. Together, these assumptions enable methods such as value decomposition [22], difference rewards [16], or centralized critics [8] to allocate credit.

## 2.3 Openness

Unlike closed systems with fixed entities, Open Agent Systems (OASYS) require agents to make decisions under structural uncertainty and compositional variability [9] because key elements such as agents, tasks, and agent types may dynamically change over time. Researchers have identified three distinct forms of openness: (1) Agent openness, where agents may leave (e.g., firefighting robots disengaging to recharge) or join dynamically; (2) Task openness, where tasks appear or disappear during operation (e.g., new passengers requesting rides in a ridesharing system); and (3) Type openness, where agent capabilities, preferences or goals evolve over time (e.g., an employee promoted to a new role with expanded responsibilities). These dynamics challenge traditional learning and planning methods and demand architectures that support flexible reasoning and adaptation in environments with changing team structure, task sets, and agent types [2].

## 2.4 MARL Settings

MARL settings are typically categorized as *competitive*, *cooperative*, or *mixed* [10, 32, 35]. In *competitive* settings, agents have opposing goals and rewards (e.g., zero-sum games), and credit assignment focuses on exploiting adversaries rather than coordinating [9, 30]. In *cooperative* settings, agents work toward a shared objective and often receive joint rewards, requiring coordination and CA among teammates [13]. *Mixed* settings involve elements of both cooperation and competition, where agents may form temporary alliances or have partially aligned objectives. These settings differ fundamentally in how agent interactions influence the learning process and the interpretation of rewards.

Among these, CA is most challenging in *cooperative* environments due to the use of shared rewards. When all agents receive the same reward, it becomes difficult to determine which agent's actions were responsible for the outcome [36]. Temporal and structural credit signals become entangled, complicating both value estimation and policy gradient updates. This ambiguity makes it harder to propagate meaningful learning signals and leads to credit misattribution. To capture and analyze these challenges, we focus on a fully cooperative environment in our empirical study (Section 4), as it provides a clear and demanding testbed for evaluating the robustness of credit assignment mechanisms under openness.

## 3. Conceptual Analysis

This section presents a conceptual analysis of how agent, task, and type openness violate the key assumptions (Section 2.2) of CA methods. Table 1 summarizes the impacts of the different aspects of openness on the key assumptions.

Each type of openness introduces distinct challenges for credit assignment. To support a more detailed analysis, we further categorize openness into specific subtypes, to differentiate between permanent and temporary changes for agent openness and task openness. Specifically, for agent openness, we consider (a) agent turnover (permanent join/leave) and (b) agent absence (temporary exit and return). Similarly, or task openness, we consider (a) task turnover (new tasks or cancellations); (b) task absence (temporary removal and reappearance). For type openness, we also consider (a) capability change; (b) preference change; (c) goal change, where each change may be temporary or permanent.

### 3.1 Agent Openness and Its Impact on CAP

Agent openness consists of agent turnover and agent absence. Agent turnover, where agents permanently join or leave the system, breaks the fixed agent set assumption and complicates credit propagation by removing contributors mid-episode. TCA cannot connect partially completed actions to future rewards, causing misattribution or loss of credit. SCA, such as centralized critics [21, 27] and value decomposition [22], must recompute reward attributions and gradient flows for a new team composition, often resulting in instability and unfairness.

| Openness Subtype | (1) Fixed Agent Set | (2) Fixed Task Set | (3) Stationary Reward Function | (4) Markov Property | (5) Action-Outcome Mapping |
|---|---|---|---|---|---|
| AO: Agent Turnover | ✘ | — | ✘ | — | ✘ |
| AO: Agent Absence | ⚠ | — | ⚠ | ⚠ | ⚠ |
| TO: Task Turnover | — | ✘ | ✘ | — | ⚠ |
| TO: Task Absence | — | ⚠ | ⚠ | — | ⚠ |
| TyO: Capability Change | ✘ | — | ⚠ | — | ⚠ |
| TyO: Preference Change | — | — | ✘ | ⚠ | ⚠ |
| TyO: Goal Change | — | ⚠ | ✘ | — | ⚠ |

**Table 1.** Impact of the different aspects of openness on key CA Assumptions (✘ breaks assumption; ⚠ complicates assumption; — no significant effect.).

Agent absence, where agents temporarily leave and later return, introduces discontinuities in state transitions and participation. During absence, expected return values and advantage estimates degrade due to missing inputs. When the agent re-enters, partial histories are incomplete, leading to gaps in reward propagation for TCA, and interrupted or inconsistent gradient updates for SCA. These changes complicate the Markov property and action-outcome mapping assumptions.

### 3.2 Task Openness and Its Impact on CAP

Task openness includes task turnover and task absence. Task turnover, where new tasks arrive or existing ones are removed, breaks the assumption of a stationary environment and stable reward function. When tasks are removed, agents who contributed toward them may receive no reward. When tasks are added, the credit landscape changes, and agents not responsible for the new tasks may still be penalized. These dynamics distort credit propagation paths in both temporal and structural methods.

Task absence, where tasks temporarily disappear and later return, introduces gaps between actions and rewards. Temporal methods lose continuity, as delayed rewards may not align with earlier actions due to intermediate task inactivity. Structural methods cannot assign credit properly to tasks that vanish before completion or reappear later, leading to missing or unfair attributions. These failures arise from violations of assumptions on consistent task sets and reward consistency.

### 3.3 Type Openness and Its Impact on CAP

Type openness includes capability change, preference change, and goal change. These changes can be temporary or permanent and may occur independently or alongside agent openness. Capability changes effectively alter the agent's functional identity, requiring reevaluation of its contributions. From a CAP perspective, this is similar to agent turnover and breaks assumptions about fixed agent roles.

Preference and goal changes shift an agent's task selection and behavior mid-episode. This alters the mapping between actions and outcomes, which violates the assumption of consistent reward functions. Temporal credit propagation is complicated when the agent's priorities shift, leading to delayed or irrelevant rewards. Structural methods struggle to fairly allocate credit when agents modify their contribution strategies. These transitions compound the challenges of agent and task openness and require CAP methods to adapt to evolving types over time.

## 4. Empirical Analysis

This section evaluates how openness impacts CAP through empirical analysis. We adopt the wildfire domain [37] and apply DQN [11] for TCA, and MAPPO [34] for SCA. We analyze how agent openness, task openness, and type openness affect credit propagation in these two methods.

We selected DQN and MAPPO because they are representative of widely used approaches in RL, respectively. DQN is a foundational method for TCA that uses value-based updates and is known to struggle in non-stationary settings without explicit adaptation mechanisms. MAPPO, on the other hand, is a state-of-the-art method for SCA in cooperative MARL that uses centralized critics to estimate joint advantages, making it highly sensitive to structural changes configurations. These characteristics make DQN and MAPPO appropriate choices for studying the impact of openness on TCA and SCA.

### 4.1 Experiments Setup

For our experiments, we used the wildfire domain [37] which supports static and open configurations. We trained the models using the official wildfire configuration WS1 under various openness conditions: no openness, agent openness, task openness, type openness, and full openness (i.e., all three types of openness combined). WS1 is from MOASEI competition for OASYS [14, 37]. WS1 configuration is a 3x3 grid, all agents present and two medium fires (intensity 2) exist. Agent openness is modeled by the need for recharging the suppressant or repairing the damaged equipment, during this process agents cannot fight a fire, thus they can do a NOOP action. Task openness is modeled by change in intensity of existing fires, either decrease by firefighters, or increase by NOOP, also there is new fire creation over time. Type openness is modeled by changing the equipment, when equipment degrades or repaired, agents range, capacity and fighting power, change respectively. When there is no fire left, the environment ends, thus the maximum step for each episode is relative to actions and events during the episode. Note that in the wildfire domain provided by MOASEI, agents do not move, actions are either fight/suppress (0) or NOOP (-1).

In our experiments, we apply padding to accommodate variable observation sizes and use action masking to manage changes in the action space. Padding inserts placeholder values to maintain a consistent observation shape when the number of observed agents or tasks varies. Action masking disables invalid or unavailable actions in dynamic settings by preventing the policy from selecting them. These techniques are necessary because, to the best of our knowledge, no existing implementation of DQN or MAPPO can handle truly unbounded observation or action spaces. While this approach enables partial modeling of openness, it does not capture its full depth or complexity.

To ensure implementation correctness, we validated our models using the PistonBall domain from PettingZoo [26], a standard MARL benchmark. The results were consistent with established baselines. Code is available at: *https://github.com/bboyfury/openness_CAP*

Additionally, we clarify that in the literature, the term "dynamic" is often used to describe changes in the environment that correspond to bounded aspects of openness as defined in Section 2.

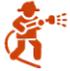

**Figure 1**. WS1 configuration where all agents present; two medium fires; slow new fire creation, with base fire spread rate 0.1 and random ignition probability of 0.05.

### 4.2 DQN

DQN assign credit temporally by learning values for state-action pairs. In open environments, however, DQN's TCA degrades. When agents enter/leave or tasks change, DQN's value estimates become outdated. For example, Pettet et al. (2023) found that as the environment transitions changed (e.g. dynamics like CartPole's pole mass), a DQN's rewards collapsed to near-zero [15].

Likewise, in a continual-learning benchmark (Agar.io), introducing ever-changing dynamics cut DQN's performance to about half of its static-task performance [12]. DQN struggles to tell whether a drop in reward is due to its own actions or simply a shift in the task or other agents' presence. In effect, credit may be mis-assigned to the wrong action when environment changes are not recognized. The result is slow or unstable learning: agents need much more experience to recover optimal behavior, and previously learned policies often become useless once the context changes.

### 4.3 MAPPO

MAPPO [34] is a widely used algorithm based on the Centralized Training with Decentralized Execution (CTDE) paradigm [34]. In CTDE, agents are trained with access to global state information and centralized critics, enabling more accurate estimation of joint value functions and advantage terms. However, at execution time, each agent acts based only on its local observations, preserving decentralization. While this setup helps better coordination and learning stability in closed systems, it implicitly relies on stable team structures and consistent joint representations during training.

In open multiagent settings, MAPPO likewise suffers CAP. Empirical studies show that multi-agent policies degrade sharply when agents change over time: for instance, in a cooperative coverage task, removing just one agent increased task completion time by ~50%, and removing two nearly tripled it [29]. This indicates that MAPPO's learned coordination fails when team structure changes. MAPPO learns to allocate credit based on a fixed set of tasks and interactions. When tasks change on-the-fly, MAPPO must effectively relearn which agent contributed to what outcome.

Even advanced MAPPO variants acknowledge these limitations: recent work notes that standard MAPPO "struggles with credit assignment" as team size or complexity grows, motivating new algorithms (e.g. PRD-MAPPO) to dynamically partition credit among subgroups [7]. If an agent's type changes, MAPPO's policy cannot automatically reassign the old agent's tasks to others. The literature suggests that MAPPO, like DQN, tends to treat each agent's policy separately and lacks flexibility for dynamic grouping, leading to misattributed rewards when types change.

### 4.4 Results

We trained DQN and MAPPO for 160,000 episodes in the wildfire domain under various openness conditions, ensuring convergence based on stabilized rewards and decreasing loss trends. After training, we evaluated the learned policies across 250 independent runs with a fixed random seed (42) to assess generalization and stability.

***Overall Performance: Average Episode Reward.*** Figure 2 illustrates the effect of different forms of openness on average episode reward, serving as an empirical signal of CA performance. In the *No Openness* setting, both methods reach their highest rewards, which confirms that all underlying CAP assumptions hold. This validates that in static environments, both temporal and structural credit can be reliably assigned, supporting consistent learning and coordinated agent behavior by achieving the high reward.

In the *Agent Openness* condition, reward drops significantly. This aligns with our conceptual analysis: agent turnover and absence violate the fixed-agent set assumption and introduce missing information in action-outcome chains. This uncertainty reduces the reliability of advantage estimates or value targets. As a result, agents learn to act more independently and less cooperatively. In the wildfire domain, this means fires are more likely to burn out due to under-coordination, leading to missed suppression opportunities and increased -1 penalties for unextinguished fires.

*Task Openness* further reduces reward. This is consistent with our observation that task turnover and task absence break the assumptions of reward function stability and action-outcome consistency. When tasks disappear or appear mid-episode, agents may not receive credit for actions toward unresolved tasks, and new task dynamics confound the propagation of responsibility. Agents cannot correctly anticipate or share responsibility, reducing their ability to coordinate. The resulting loss in cooperative coverage again increases fire spread and penalties, diminishing the average reward.

In the *Type Openness* setting, the decline in performance is less severe but still notable. Changes in agent capabilities, preferences, or goals mid-episode do not remove agents but instead alter the context of their contributions. This violates assumptions about stable agent types and introduces uncertainty in interpreting an agent's intent and impact. As a result, other agents can no longer infer reliable coordination patterns, and CAP mechanisms struggle to differentiate between intentional contributions and behavior change. In the wildfire domain, this misalignment can lead to partial suppression or underpowered actions, increasing the likelihood of unresolved fires and associated penalties.

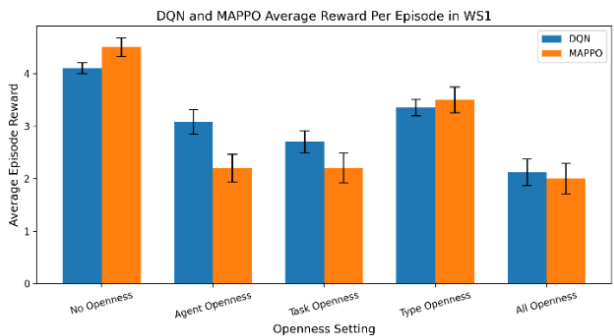

**Figure 2.** Average episode reward under different openness settings. Openness consistently degrades reward due to violations of assumptions required for accurate credit propagation. Agent and task dynamics have particularly strong effects, while combined openness leads to the most severe CAP failure.

The most severe degradation occurs under All Openness, where agent, task, and type change over time. This setting simultaneously violates all five key assumptions for reliable CA. Both temporal and structural signals become inconsistent and noisy, leaving agents unable to learn stable policies or trust their teammates' behavior. Coordination collapses, and agents resort to myopic or unaligned actions. As a result, large portions of the grid go unprotected, fires burn out uncontrollably, and the system accumulates heavy -1 penalties due to task failure.

*Impacts of Openness on TCA.* Figure 3 Shows that in the *No Openness* condition, DQN loss decreases smoothly and steadily throughout training. This behavior is expected, as all temporal dependencies remain intact and the reward function is stable. The network consistently receives valid updates through bootstrapped targets, enabling clean convergence of Q-values. This confirms that when assumptions of stationarity and fixed agent/task sets are preserved, TCA operates as intended.

Under *Agent Openness*, *Task Openness*, and especially *All Openness*, DQN exhibits significant variance in the loss curve. Loss remains elevated and volatile for a large portion of training, with slower and noisier descent. The high variance in DQN loss indicates unstable TD-errors. This is symptomatic of TCA failure, as the agent cannot form a stable value estimate for its actions when the connection between an action and its delayed reward is broken by agents leaving or tasks changing.

The instability seen in the *All Openness* condition demonstrates how compound violations, dynamic tasks, agents, and capabilities, cause bootstrapped estimates to become misleading. DQN cannot differentiate whether loss stems from a bad policy or an altered environment. This confounds learning signals and results in frequent Q-value oscillations, which are evident in the persistent fluctuations throughout training.

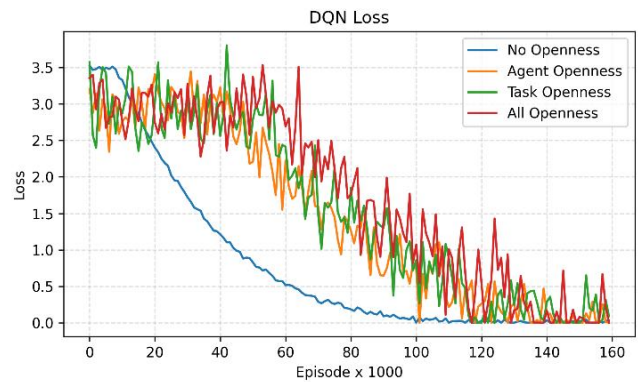

**Figure 3.** DQN Loss Under Openness. Loss declines smoothly in static settings but shows high variance and delayed convergence under openness. This reflects instability in TCA as key assumptions are violated.

*Impacts of Openness on SCA.* Figures 4 and 5 Indicate that in the No Openness setting, both the actor and critic losses of MAPPO steadily decrease, with minimal variance. This behavior reflects the effectiveness of SCA when the environment remains fixed. The centralized critic can consistently estimate joint advantages, and the actor receives coherent gradients to optimize policy updates.

In contrast, under all openness conditions, both actor and critic losses display increased variance and slower convergence. In particular, Agent Openness and All Openness lead to erratic updates, indicating instability in estimating advantages and returns. The noisy actor-critic losses (Figures 4 and 5) demonstrate the centralized critic's inability to produce a consistent advantage function. This is an indication of SCA failure, as the critic can no longer correctly attribute a team's success or failure to individual agents when the team's structure is in flux.

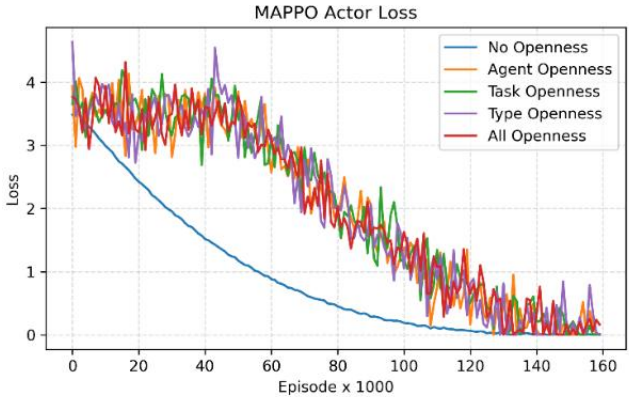

**Figure 4.** MAPPO Actor Loss Under Openness. Actor loss decreases smoothly under No Openness but becomes volatile with openness conditions. Agent, task, and type openness introduce structural instability, resulting in noisy policy updates and delayed convergence. This reflects weakened structural credit assignment due to dynamic agent-task relationships and complicated gradient flow.

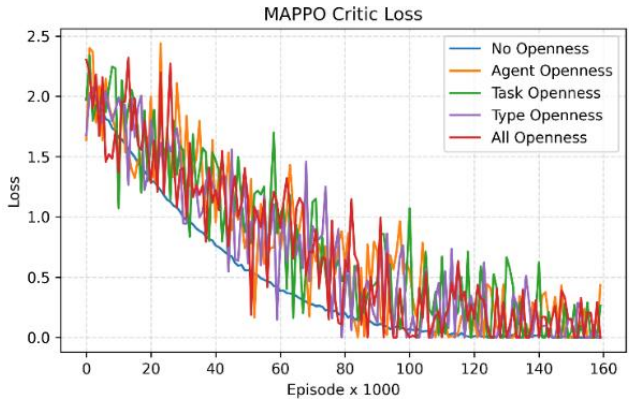

**Figure 5.** MAPPO Critic Loss Under Openness. Critic loss under No Openness shows stable convergence, while all forms of openness induce higher variance and slower loss decay. The centralized critic fails to produce consistent value estimates when the environment structure changes, indicating impaired credit attribution across evolving agent and task compositions.

The Task Openness condition introduces slightly less instability than Agent Openness or All Openness, but the loss curves are still noisier compared to the baseline. Task appearance and disappearance violate the assumptions of a fixed task set, stationary reward function, and consistent action-outcome mapping, which affect the centralized critic's ability to compute advantage estimates across agents, resulting in higher variance in both actor and critic loss. Meanwhile, Type Openness introduces moderate noise, as type changes break the mapping between agent actions and their intended outcomes, thereby complicating both value estimates and policy gradients.

It is also worth mentioning that during training we observed under openness, other than noisy loss and high variance in the reward, convergence took longer than non-open configuration, i.e., around 1.2 times more.

*Unboundedness*. We emphasize that, in our implementation, we applied observation padding and action masking to convert the inherently unbounded nature of openness into a bounded problem. This design choice allows traditional algorithms like DQN and MAPPO, which are not inherently equipped to handle unbounded observation or action spaces, to operate in open environments. However, this is a workaround rather than a solution.

True unboundedness, as introduced by openness, remains a fundamental challenge. We expect that in fully unbounded settings, where the number of agents, tasks, or types is not artificially constrained, traditional methods would experience significantly worse performance due to their inability to generalize CA under such dynamic conditions.

## 5. Research Directions Analysis

This section outlines gaps in existing CAP methods for MARL under openness, highlighting possible directions for future research. Current literature and our empirical analyses indicate significant shortcomings in handling agent openness, task openness, and type openness, suggesting opportunities for the future advancements.

A significant gap in TCA is the reliance on assumptions of stable environmental dynamics and consistent reward propagation pathways, both complicated by openness. Traditional TCA methods, such as temporal difference (TD) learning [27] and eligibility traces [21], fail to reliably connect actions to delayed rewards when agents enter or leave, or tasks dynamically change. To consider openness, future research should investigate TCA mechanisms that explicitly accommodate environmental non-stationarity by dynamically recalibrating credit propagation paths. Promising approaches include adaptive eligibility traces [21], dynamically adjusted discount factors [1], and context-sensitive bootstrapping techniques [33].

For SCA, the key research gap lies in the inability of current methods to dynamically attribute outcomes to changing sets of agents or tasks in the policy gradient methods. Networks and fixed decomposition methods typically assume stable agent compositions and task structures, leading to credit misattribution under agent turnover, absence, or type changes. Future work should explore flexible SCA frameworks that inherently adapt their credit attribution strategies as structural dynamics evolve. Potential methodologies include dynamic graph-based methods [18], attention mechanisms for selective attribution [3, 28], and adaptive decomposition approaches [22], that realign CA structures in real-time.

Finally, there is a crucial evaluation gap concerning openness-aware CAP methods. Existing benchmarks typically reflect simplified or artificially bounded openness scenarios, limiting the validity of CAP evaluations. Establishing rigorous, realistic benchmarks that accurately represent complex openness scenarios in OASYS is vital for validating and refining openness-resilient CAP methods comprehensively.

## 6. Conclusion

This work analyzed the impact of openness on the CAP in MARL. Through thorough conceptual and empirical analyses, we demonstrated how agent openness, task openness, and type openness complicate both TCA and SCA. Our analysis highlighted critical violations of foundational CAP assumptions, including stationary environments, fixed agent compositions, stable reward functions, the Markov property, and consistent action-outcome mappings.

Empirical evaluations using representative CAP approaches revealed degradation in CA effectiveness under openness conditions, resulting in increased instability, delayed convergence, and reduced overall coordination. Combined openness conditions particularly exacerbated these impacts, severely compromising system performance.

Overall, this study contributes conceptual and empirical perspectives on how openness challenges credit assignment. It establishes sub-categories of openness events, demonstrates their disruptive effects on representative temporal and structural algorithms, and defines concrete research needs for openness-robust CAP frameworks. We identified several focused research gaps that must be addressed: developing adaptive TCA mechanisms, flexible SCA frameworks, robust credit attribution methods capable of handling agent type variability, and inherent methods for managing unbounded openness. Addressing these targeted gaps will significantly enhance the effectiveness of CAP methods, making MARL systems more resilient and reliable in realistic, open-agent scenarios.

## Acknowledgements

This research was supported, in part, by a collaborative NSF #IIS-2312658. Computing occurred on the Holland Computing Center of the University of Nebraska, which receives support from the university's Office of Research and Economic Development and the Nebraska Research Initiative.